\documentclass[fleqn,10pt]{article}
\usepackage[utf8]{inputenc}
\usepackage[T1]{fontenc}
\usepackage{subcaption}
\usepackage{floatrow}
\usepackage{graphicx}
\usepackage{soul}

\newfloatcommand{capbtabbox}{table}[][0.45\textwidth]

\title{Automatic Posture and Movement Tracking of Infants with Wearable Movement Sensors}
\date{}
\author{Manu Airaksinen$^{1,2}$, Okko R\"{a}s\"{a}nen$^{1,3}$, Elina Il\'{e}n$^{4}$, Taru H\"{a}yrinen$^2$,\\ Anna Kivi$^2$, Viviana Marchi$^{5,6}$, Anastasia Gallen$^2$, Sonja Blom$^2$, \\
Anni Varhe$^2$, Nico Kaartinen$^7$, Leena Haataja$^2$, Sampsa Vanhatalo$^{2,8,9}$}

\begin{document}

\maketitle
\noindent $^1$Department of Signal Processing and Acoustics, Aalto University, Espoo, Finland \\
$^2$BABA center, Pediatric Research Center, Children's Hospital, Helsinki University Hospital and University of Helsinki, Finland \\
$^3$Faculty of Information Technology and Communication Sciences, Tampere University, Tampere, Finland \\
$^4$Department of Design, Aalto University, Espoo, Finland \\
$^5$Institute of Life Sciences, Scuola Superiore San'Anna and Department of Developmental Neuroscience, IRCCS Fondazione Stella Maris, Pisa, Italy \\
$^6$Department of Developmental Neuroscience, IRCCS Fondazione Stella Maris, Pisa, Italy \\
$^7$Kaasa solution GmbH, D\"{u}sseldorf, Germany \\
$^8$Department of Clinical Neurophysiology, HUS Medical Imaging Center,  University of Helsinki, Helsinki University Hospital and University of Helsinki, Finland \\
$^9$Neuroscience Center, University of Helsinki, Helsinki, Finland \\
\begin{abstract}
Infants' spontaneous and voluntary movements mirror developmental integrity of brain networks since they require coordinated activation of multiple sites in the central nervous system. Accordingly, early detection of infants with atypical motor development holds promise for recognizing those infants who are at risk for a wide range of neurodevelopmental disorders (e.g., cerebral palsy, autism spectrum disorders). Previously, novel wearable technology has shown promise for offering efficient, scalable and automated methods for movement assessment in adults.
Here, we describe the development of an infant wearable, a multi-sensor smart jumpsuit that allows mobile accelerometer and gyroscope data collection during movements. Using this suit, we first recorded play sessions of 22 typically developing infants of approximately 7 months of age. These data were manually annotated for infant posture and movement based on video recordings of the sessions, and using a novel annotation scheme specifically designed to assess the overall movement pattern of infants in the given age group. A machine learning algorithm, based on deep convolutional neural networks (CNNs) was then trained for automatic detection of posture and movement classes using the data and annotations. Our experiments show that the setup can be used for quantitative tracking of infant movement activities with a human equivalent accuracy, i.e., it meets the human inter-rater agreement levels in infant posture and movement classification. We also quantify the ambiguity of human observers in analyzing infant movements, and propose a method for utilizing this uncertainty for performance improvements in training of the automated classifier. Comparison of different sensor configurations also shows that four-limb recording leads to the best performance in posture and movement classification.
\end{abstract}
\flushbottom
\maketitle

\thispagestyle{empty}

\section*{Introduction}
A key global healthcare challenge is the early recognition of infants that eventually develop lifelong neurocognitive disabilities. More than every tenth infant is considered to be at neurodevelopmental risk  \cite{Rosenberg2008}
due to their neonatal medical adversities, such as prematurity, birth asphyxia, stroke, metabolic derangements and intrauterine substance exposures. An early therapeutic intervention would be optimal for reducing the ensuing lifelong toll to individuals and societies, although it has been challenging to efficiently target early intervention to those who actually would benefit from it \cite{Novak2017}.
A thorough screening of all risk infants with extensive test batteries and/or brain imaging techniques is not plausible in most parts of the world, and it is not justifiable even in the most developed and wealthy nations.

 There is hence a rising need to  develop generalizable, scalable, objective, and effective solutions for early neurodevelopmental screening. Such methods would need to be robust to random variability that may arise from all parties involved: the infant him/herself, skills of the health care professional, testing environment, as well as recording methods. Furthermore, a wide biological variation which is inherent to typical neurodevelopment has to be considered. Indeed, requirement for adequate robustness has been a key challenge in the recent work \cite{Lobo2019}.
 
Phenomenologically, assessment of infants' spontaneous behavior has recently gained a lot of interest for three reasons: First, the lab or hospital environments are artificial from a child's perspective, and hence challenging for assessing their true neurodevelopmental performance \cite{Nazneen2015}. Second, the novel recording and analysis methodologies have made it possible to study participants' spontaneous activities in both lab and other environments \cite{Rihar2014, Venek2018}
Third, both fundamental and applied research has shown that infant's spontaneous movements may provide an important global window to the infant brain function \cite{Einspieler2008,Peyton2017}.
Besides bodily movements \emph{per se}, development of control of posture and intentional voluntary movements can be seen as parallel processes forming one perceptual-motor system which also involves higher cognitive functions \cite{Harbourne2015}. Characterization of infants' typical pattern of variation in different postures and movement activity over longer time could be used as a tool for early screening of infants at neurodevelopmental risks. Ideally, such a system would consist of an easy-to-use recording setup applicable to home environments, followed by an automated analysis pipeline for objective and quantitative assessment. Widely used observation protocols have been developed for assessment of spontaneous movements in neonatal period and early infancy \cite{Novak2017, Guzzetta2003}. However, they are neither genuinely quantitative nor allow longitudinal tracking beyond four months of age.

One option for monitoring infant movements at their homes could consist of intelligent wearables with integrated sensors. There has been significant recent progress in the development of intelligent wearables for sports and leisure clothing in adults \cite{Peake2018}. However, we are not aware of standards or open solutions for multi-sensor-based movement analysis for infants. Our present work aims to fill this gap by describing a relatively inexpensive comfortable-to-wear, and easy-to-use intelligent wearable, a \emph{smart jumpsuit} that can be used for monitoring and quantifying key postures and movement patterns of independently moving infants. In addition to the design of the jumpsuit itself, we have developed a new protocol to visually classify and annotate independent movements of infants yet to acquire upright posture. We also developed a machine learning -based classifier to automatically recognize the set of postures and movements covered by the annotation scheme by using the sensory data available from the jumpsuit sensors, including a novel way to deal with inter-annotator inconsistencies inherently present in the human annotations used to train the classifier. Performance of the resulting system was assessed against multiple human raters on data from infants previously unseen by the classifier. The results show that the system achieves posture and movement recognition accuracy comparable to human raters on the same data, indicating the feasibility of automatic assessment of spontaneous infant movements using the proposed smart jumpsuit.

\begin{figure}[t]
    \centering
    \begin{subfigure}[c]{0.4\linewidth}
        \centering
        \includegraphics[width=0.75\linewidth]{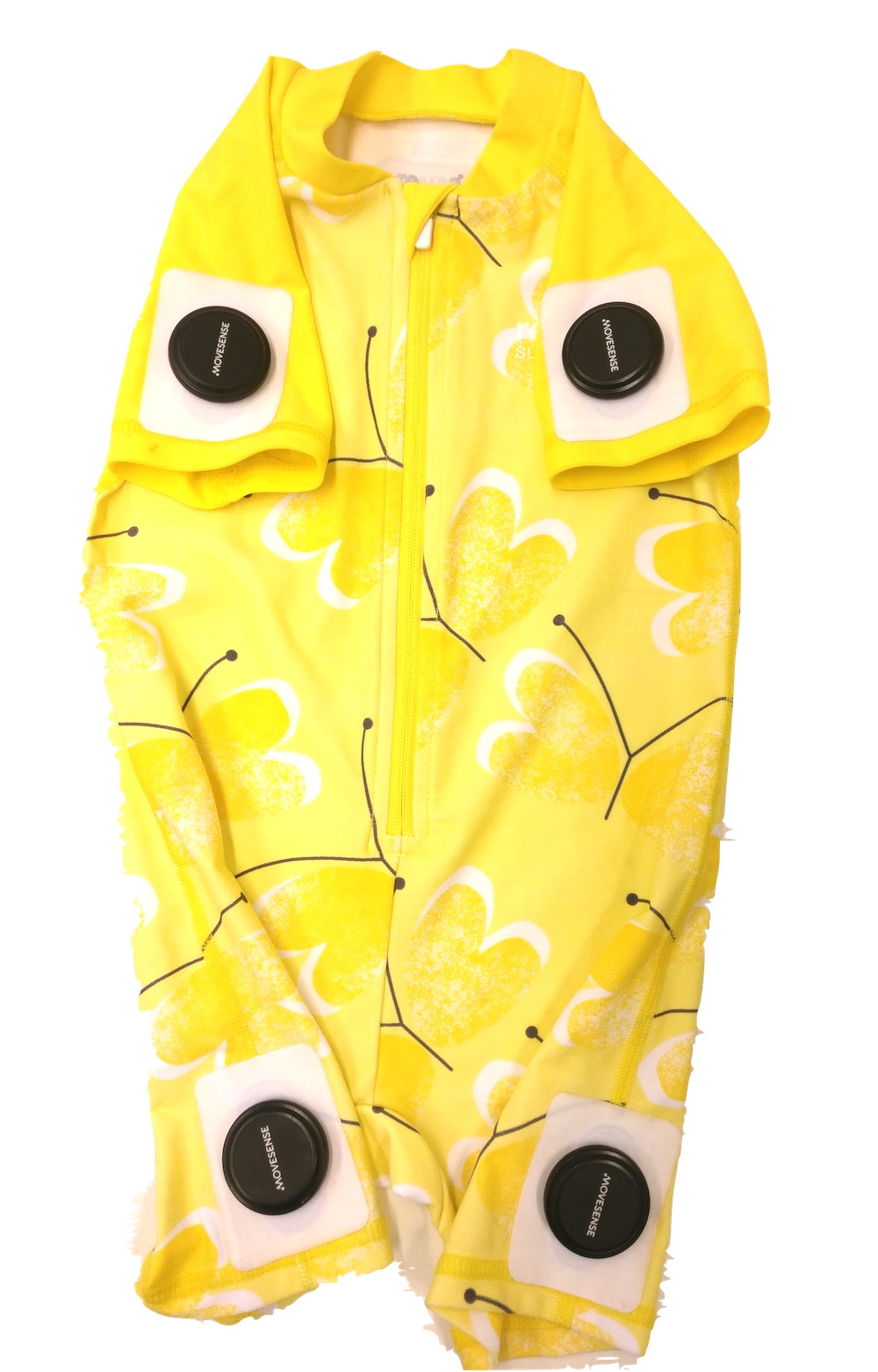}
        \caption{}
    \end{subfigure}
    \qquad 
    \begin{subfigure}[c]{0.4\linewidth}
        \centering
        \includegraphics[width=0.75\linewidth]{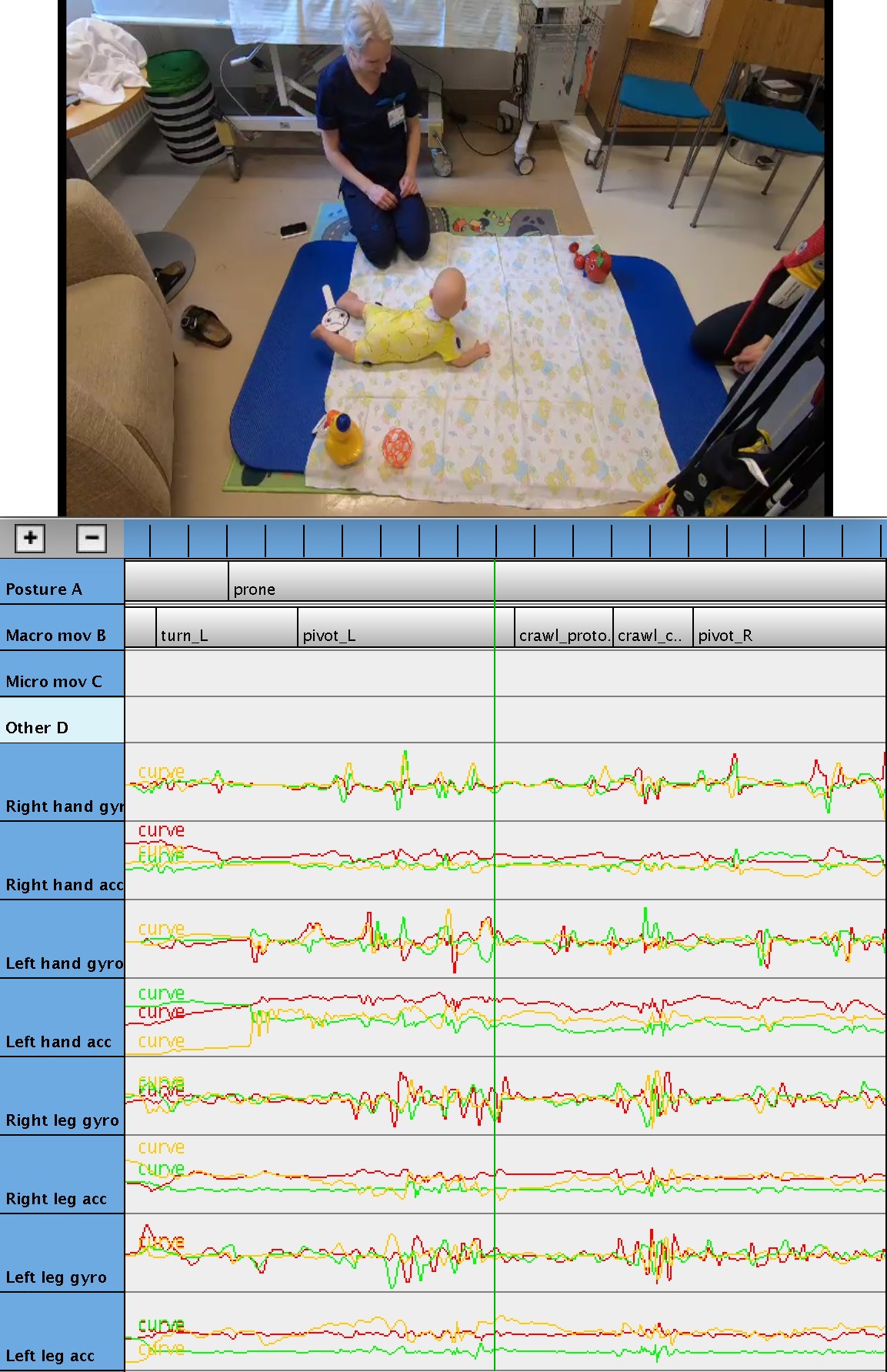}
        \caption{}
    \end{subfigure}
    \caption{Experimental design. (a) Photograph of the smart jumpsuit with four proximally placed movement sensors.  (b) The annotation setup displaying annotations with synchronized video and movement data.}\label{fig:fig1}
\end{figure}

\section*{Data collection methods}

The overall workflow in our study included i) the development of a wearable garment and mobile data collection system for infant recordings (Fig. \ref{fig:fig1}), ii) the development and implementation of the visual analysis scheme to obtain a human benchmark, and iii) the development, training, and performance testing of the machine learning methods for an automated quantitative analysis of infants' movement activity (see Fig. \ref{fig:neural_net}).

\subsection*{Design of the smart jumpsuit}
For tracking the posture and movement of infants, we developed a measurement suit, named the ``smart jumpsuit'' (see Fig. \ref{fig:fig1}a), which is a full body garment that allows spontaneous unrestricted movements. The garment features a total of four battery operated wireless Suunto Movesense sensors (www.movesense.com) that are mounted proximally in the upper arms and legs.
For the first prototype, we used a commercially available infant swimming suit as the base garment. Movesense Smart Connector mounts were set onto fabric and attached by heat bonding the piece of fabric and thermoplastic polyurethane (TPU) adhesive film-layer structure on top of the mounts. The created electronic garment is laundry washable at 40 $^{\circ}$C.

Each of the Movesense sensors features a built-in inertial measurement unit (IMU) that is used for six degrees of freedom (DOF) measurement at a sampling rate of 52 Hz. These signals consists of a triaxial accelerometer measuring linear acceleration in $m/^s2$ (range $\pm 8$g) and a gyroscope measuring angular velocity in $^{\circ}/s$ (range $\pm 500 ^{\circ}/s$). The Movesense sensors are 36.6 mm in diameter and 10.6 mm in thickness, weigh 30 grams, and are waterproof and removable from programmable mounts, making them suitable for limb placements in infant wearables. Symmetric sensor placement was used to capture a comprehensive picture of infant motor repertoire. In addition, the setup enables later more detailed study of limb movement synchrony and symmetry that can be used to further characterize clinically abnormal neurological development \cite{Guzzetta2003}.

Communication of Movesense sensors uses an open source application programming interface (API; bitbucket.org/suunto/movesense-docs/wiki/Home), which allows streaming the raw sensor data at desired frequencies wirelessly via a Bluetooth 4.0 connection to an external data logging device. This project used Apple iPhone SE with an iOS-based multi-sensor data logging software (www.kaasa.com). The tasks of the data logger software include 1) centrally controlling the recording process, 2) receiving and writing raw signals from each sensor through the Bluetooth connection, and 3) synchronizing sensor timestamps. In addition, the software allows concurrent video recordings with the device camera.

As presented, the smart jumpsuit is a technical proof-of-concept, and it is not meant to showcase a ready commercial product. The price for the present smart jumpsuit construct from retail components is approximately 200 USD, and the iOS-based mobile phone could be replaced with any of the now ubiquitous smartphones able to run a multisensor datalogger software. A larger scale production price of the smart jumpsuit garment could readily go down to 30 -- 50 USD, making it financially feasible for a wide range of use-contexts, including developing countries.

\subsection*{Data collection}
The main goal of our data collection process was to obtain a representative set of independent movements in approximately 7-month old infants for automatic classifier training. For this purpose, a total of 24 infants were recorded (mean age 6.7 months, $\pm 0.84$ (SD), range 4.5--7.7; 9 male). The infants were known as typically developing without prior history of significant medical issues, and they were initially recruited for a larger ongoing research project. The measurements used in this particular study were carried in clinic-like settings in approximately 30--60 minute long sessions. Notably, the length of recording was limited by situational factors arising in the laboratory context, such as lab availability, or infant care routines and cooperation. The recording technology (i.e., smart jumpsuit with data logger) would have allowed considerably longer recordings.

During each session, an infant wearing the smart jumpsuit was placed on a foam mattress. Without making physical contact with the infant, a pediatric physiotherapist actively engaged the infant in different postures and movements common in structured neurological examinations \cite{Haataja1999} with a set of age-appropriate toys. The toys were also freely available for the infant to play with. The caregiver of the infant was allowed to be in close proximity (if they so desired) in order to keep the infant more content. In order to ensure maximum amount of independent movement, physical contact with the infant was limited to situations that required soothing or lifting the infant in prone/supine posture if the ability to turn around was not yet acquired.

All sessions were also video recorded to allow detailed temporal annotation of movement activities (see Movement annotations Section). The first 14 infants were video-recorded using a separate video camera (GoPro Hero 8 Black, www.gopro.com), while the rest were recorded with an in-built synchronized video feature of the data-logger software on the iPhone that automatically synchronized with the recorded sensor data. For the first 14 recordings with the GoPro camera, synchronization between the sensor signals and video was done by tapping one of the sensors visibly in the video to allow manual adjusting between movement signals and video when importing into Anvil annotation software. Two infants were excluded as outliers from further analysis due to their substantially more advanced motor skills (e.g., standing), yielding a final number of 22 recordings. Each recording consists of 24-channels of sensory data (4 sensors, 3 accelerometer and 3 gyroscope channels per sensor) sampled at the 52 Hz sampling frequency. The total length of the recordings was 12.1 hours, out of which 10.6 hours (88\%) were utilized for the classifier training and testing described in the automatic detection of infant posture and movement section (excluding carrying, out-of-camera events, and dropped sensor connections). The resulting mean length of the utilized recordings per infant was 29 minutes (range 9--40 minutes).

The study complies with the Declaration of Helsinki and the research was approved by the Ethics Committee of Children's Hospital, Helsinki University Hospital. A written informed consent was obtained from the parents.


\subsection*{Movement annotation protocol}
\label{sec:annotation}
The primary goal of the present jumpsuit setup and analysis was to obtain a temporally rich description of the infant movement activities over periods of time. This information could be used later to support a variety of clinical goals and decision making. In addition, an automated classifier for such movements could be then fine-tuned or adapted for more specialized clinical diagnostics or evaluation of intervention efficacy --- tasks for which less training data are typically available compared to the overall infant population available for data collection.

Notably, infant movement patterns differ substantially from the prototypical adult categories (e.g., walking or running), and there are no standard classification systems for the recognition of infant movements in a comparable manner. The existing schemes for infants are aimed to visually recognize more complex entities via Gestalt perception, typically handled in the framework of ``movement behavior'' or ``kinematics'' \cite{Einspieler1997}. Therefore, we first developed a novel classification and annotation scheme that describes the expected typical postures and movements of three to seven months old infants before they reach upright posture \cite{Piper1994}.
The scheme was developed using an iterative approach within a multidisciplinary team of clinical and machine learning experts, as well as by comparing consensus and blinded annotations carried out on a set of pilot recordings. 
The overall aim was to reach an annotation scheme where each category would have an unequivocal verbal description, minimal overlap with other categories, maximum pervasiveness, as well as minimal disagreement in ratings from multiple annotators. The distal fine motor movements were omitted from the annotation scheme because their identification was not consistent at the desired temporal resolution. The final annotation scheme consisted of two annotation tracks, one to represent posture, and the other to represent gross body movements. This annotation design allows tracking of movements in the context of posture, as well as their further quantification (such as amplitudes, synchronies or temporal correlations) in a context-dependent manner (see, e.g., \cite{Harbourne2015}).

The posture track consists of five discrete categories: \texttt{prone}, \texttt{supine}, \texttt{side left (L)}, \texttt{side right (R)}, and \texttt{crawl posture}. The eight categories for the movement track are: \texttt{macro still}, \texttt{turn left (L)}, \texttt{turn right (R)}, \texttt{pivot left (L)}, \texttt{pivot right (R)},  \texttt{crawl proto}, \texttt{crawl commando} and \texttt{crawl 4 limbs}. The exact verbal descriptions of these event types as presented to human annotators are shown in the Supplementary Information. However, because only one of the utilized recordings contained the movement category \texttt{crawl 4 limbs}, it was omitted from further analysis in this study, reducing the number of considered movement categories to seven. The overall rationale of this particular annotation template was to identify posture/movement patterns that are elementary enough for serving as a basis for further (posture) context-sensitive, quantitative (movement) summaries; also, they should be reliable to detect with proximal movement sensors. Therefore, we deliberately chose not to aim at 
direct classification of complex motility patterns/kinematics that are commonly observed in the established observational assessment scales \cite{Novak2017}.

Each recording was independently annotated for posture and movement by three of the authors (M.A., A.K., A.G., S.B, A.V. took part in the annotations), as well as for a meta-data track for additional information, such as epochs when the infant was out of video or carried by an adult.  The use of three parallel annotators was chosen as a compromise between the need to validate the consistency of the proposed annotation scheme and the laborious effort required in the rigorous and detailed annotation task \cite{Tapani2019}. The annotators had varying amounts of experience with infant clinical research with backgrounds in biomedical disciplines, psychology or engineering.
The annotators were trained to carry out the movement annotation task, and they performed the actual annotation task only after they felt confident enough to do so. Their annotations were also evaluated after a few infants to ensure their consistency w.r.t. the proposed annotation protocol.

The recordings were annotated using freely available Anvil annotation software (www.anvilsoftware.com), in which the annotators had a simultaneous view of the playback video
and visualization of raw multi-channel signals while carrying out multi-track annotation of the data (see Fig. \ref{fig:fig1}b). Visually presenting the video-synchronized waveforms was found to be useful by markedly improving the temporal accuracy of annotations. The annotation templates as well as an example annotation file are provided in the Supplementary material. Consistency of the  annotators, and therefore of the entire annotation protocol, was quantified in terms of inter-rater agreement rates, as described below.

\section*{Annotation consistency analysis}

\begin{figure}[ht]
    \centering
     \includegraphics[width=\textwidth]{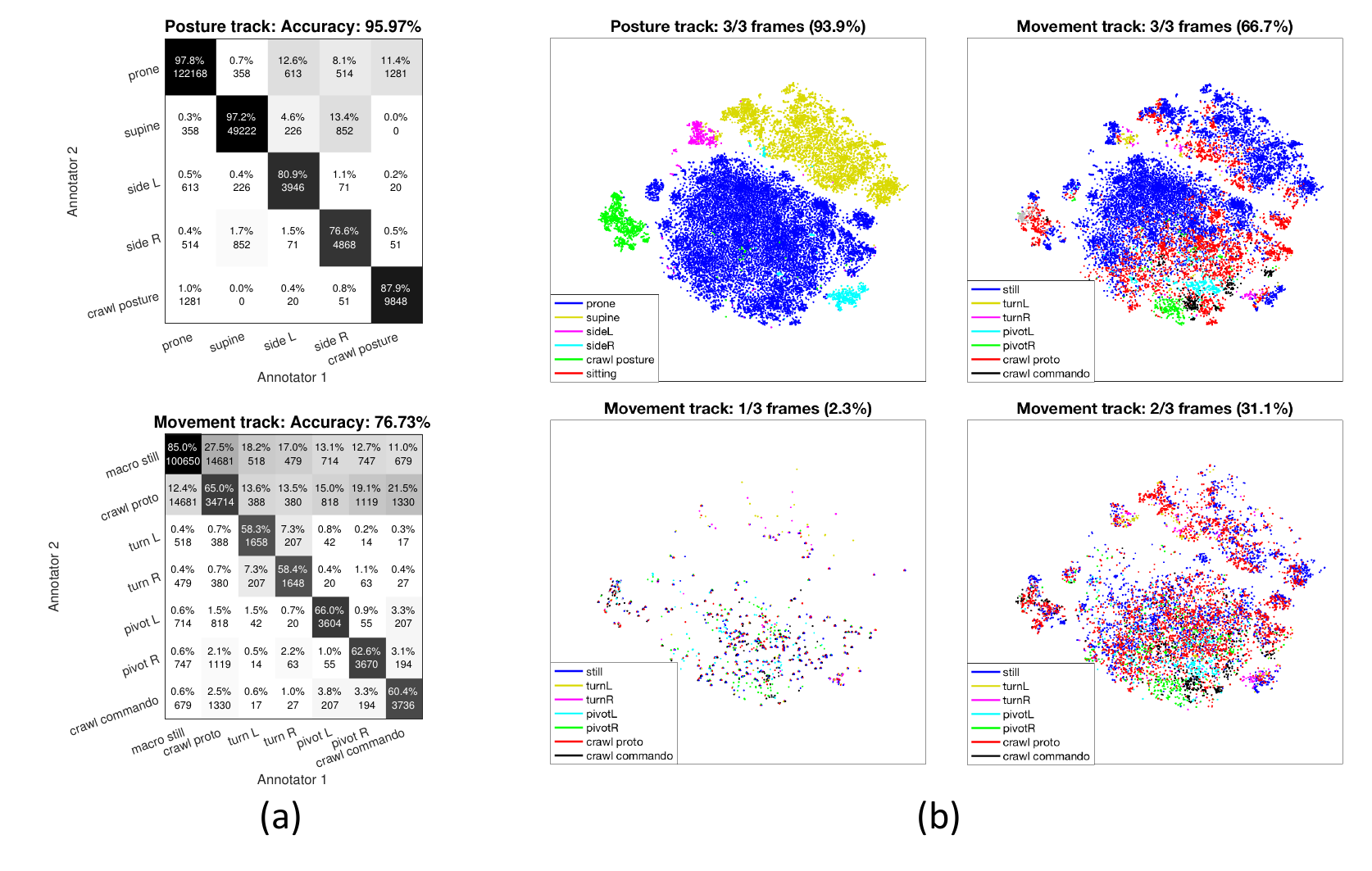}
    \caption{(a) Total cumulative confusion matrices across all possible annotator pairs for the Posture (top) and Movement (bottom) track. The percentages of each column sum up to one and the absolute values denote the number of frames corresponding each cell. (b) t-SNE visualization of the entire dataset based on SVM input features. Color coding is based on the annotations of Posture (top left) and Movement categories (the rest). The visualization of the Movement track has been broken down into 3/3 (top right), 2/3 (bottom right), and 1/3 (bottom left) annotator agreement levels. The ambiguity differences in the annotation accuracies between the tracks can be clearly seen.}\label{fig:annotaion_results}
\end{figure}

In order to evaluate the coherence of the annotations, we measured inter-rater agreement using Fleiss' $\kappa$ score (with $\kappa$ = 1 for full and $\kappa$ = 0 for chance-level agreement), which is a multi-rater generalization of the Cohen's $\kappa$ metric \cite{Cohen1960}. The overall agreement across all categories was very high for posture ($\kappa=0.923$) and moderate for movement ($\kappa=0.580$). However, breakdown of the comparisons in the confusion matrix (Fig. \ref{fig:annotaion_results}a) shows substantial variability between the classes. For the posture track, \texttt{prone} was sometimes confused with \texttt{crawl} or \texttt{side L/R} postures, and \texttt{supine} was as well confused with \texttt{side L/R} postures. For the movement track, there were several confusions with \texttt{macro still}, which could be confused with all other classes; likewise, the movement \texttt{crawl proto} was frequently confused with other movement patterns. While some incidences may be explained by human error, such as confusing left and right pivoting or turning, these disagreements between human annotators as a whole suggest substantial inherent ambiguity in movement classification, indicating the difficulty of interpreting infant movements in terms of unequivocal discrete categories.

In order to better understand the source of ambiguities in the human annotations, we plotted the entire dataset (Fig. \ref{fig:annotaion_results}b) using t-distributed stochastic neighbor embedding (t-SNE) \cite{vanDerMaaten2008} with signal features described in the SVM classifier Section below. 
In the figure, posture information has clearly separable boundaries between categories (the ``islands'' in Fig. \ref{fig:annotaion_results}b). In contrast, category boundaries between the movement classes are markedly ambiguous with a continuum of movement category (mini-)clusters inside the islands. This further illustrates how the posture classification is a relatively easy task whereas accurate movement categorization is much more challenging. 

The inherent ambiguity of certain movement categories and their temporal extent was also reported by the human annotators during feedback discussions. According to the subjective experiences of the annotators, the most typical cases of ambiguity were concerned with 1) the determination of the exact temporal boundaries between two subsequent movements, and 2) deciding when rapid transient movements reached the criteria for belonging to a movement category (see Fig. \ref{fig:annotaion_results}a for the confusion matrices). The annotators also reported the possibility of a ``recent observation effect'' \cite{Summerfield2015} in their annotations, i.e., the subjective criteria for a category (e.g., proto-crawling) could have been more loose for a generally less active infant as compared to an active infant. Overall, the analysis shows that despite the structured annotation protocol, annotation of infant movements based on visual inspection is not fully unambiguous due to the temporal contiguity of movements and postures, and due to the continuous nature of the motor development that infants are undergoing.

\section*{Automatic detection of infant posture and movement}
\subsection*{Training of automatic classifiers for posture and movement}

We explored the capabilities of the smart jumpsuit in measuring the proposed posture and movement categories in conjunction with two different classifier architectures: First, a support vector machine (SVM) classifier based on established signal-level features from the human activity detection literature \cite{Yang2008, Anguita2013}, and second, a new end-to-end convolutional neural network (CNN) architecture designed for the task at hand. While the SVM provides a baseline performance level for the given task, the feature-agnostic CNN is potentially more powerful, as it can learn task-relevant signal representations directly from the training data instead of using a prescribed set of signal features. However, the gain from an end-to-end system is known to become significant with datasets that are orders of magnitude larger than our present dataset \cite{Zeyer2018}, so we wanted to assess whether the CNN is able to provide any improvement over the well-established SVM baseline.

\subsubsection*{SVM classifier}
\label{sec:SVMmethod}
For both SVM and CNN classifiers, the signal was first windowed into 120-sample frames (2.3 s at 52 Hz) with 50\% overlap between subsequent windows. For the SVM classifier, 14 basic features per channel assembled from prior literature \cite{Yang2008, Anguita2013} were then calculated, yielding a feature vector with a total dimensionality of $14\times 24=336$ features per frame. The utilized features were: signal mean, variance, max amplitude, min amplitude, signal magnitude area, energy, interquartile range, skewness, kurtosis, largest frequency component, weighted average frequency, frequency skewness, and frequency kurtosis of each channel. The multi-class SVM was trained as an ensemble system with the error-correcting output codes (ECOC) model \cite{Allwein2001} utilizing linear kernel functions. The input vectors to the model were standardized with global mean and variance normalization.

\begin{figure}[t]
    \centering
    \begin{subfigure}[c]{0.9\linewidth}
        \centering
        \includegraphics[width=\linewidth]{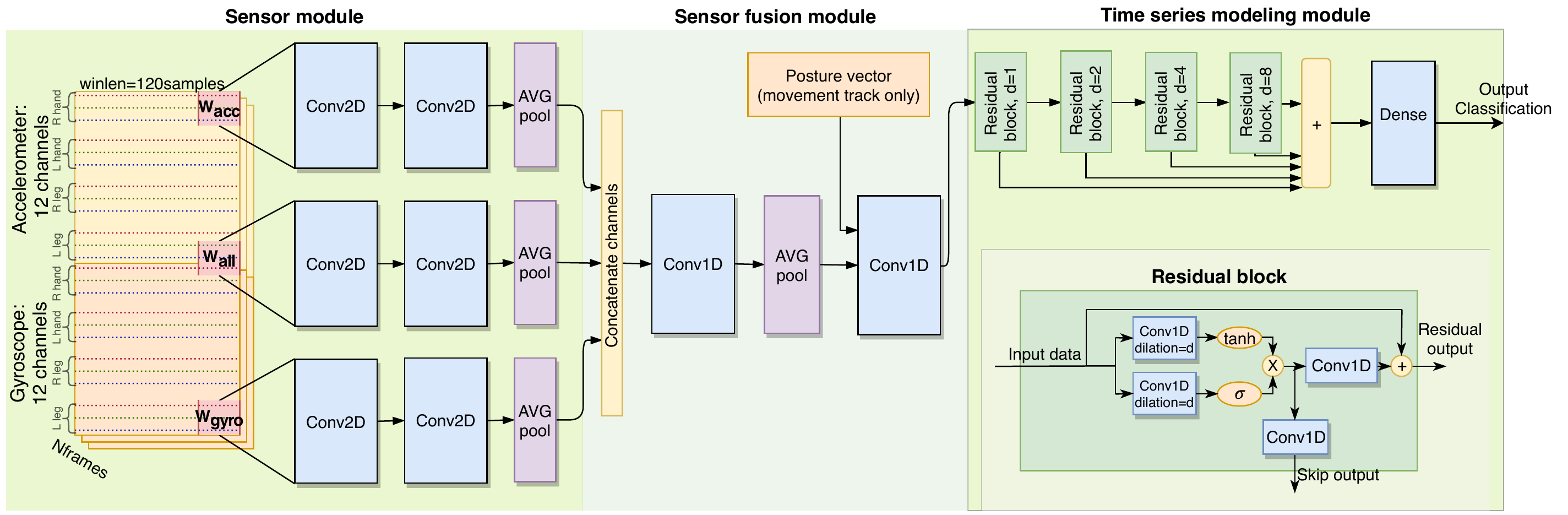}
        \caption{}
    \end{subfigure}
    \\ 
    \begin{subfigure}[c]{0.9\linewidth}
        \centering
        \includegraphics[width=\linewidth]{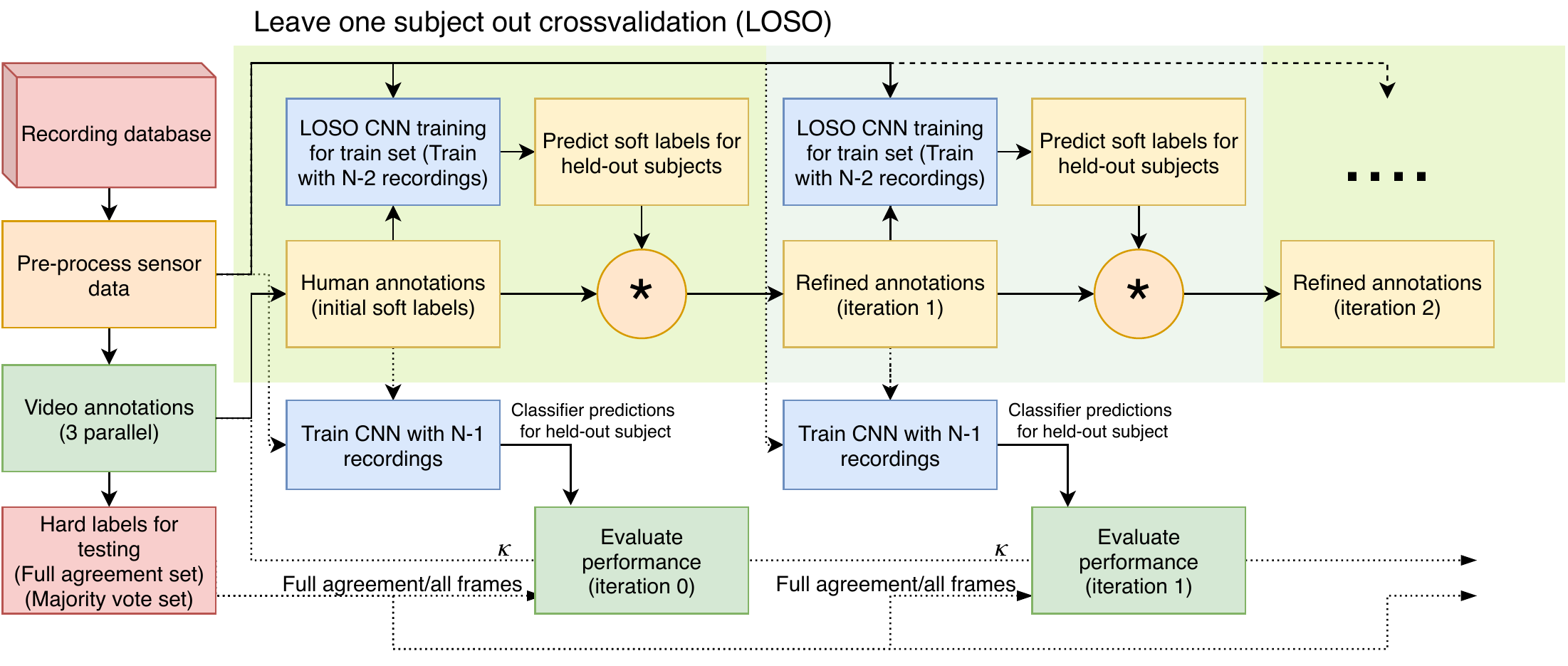}
        \caption{}
    \end{subfigure}
    \caption{Design of the automatic classification pipeline. (a) The convolutional neural network (CNN) architecture used in the study as the main classifier. The role of the sensor module is to perform sensor-specific feature extraction, the sensor fusion module fuses sensor-level features into frame-level features, and the time series modeling module captures the temporal dependencies across frame-level features. (b) Block diagram for the iterative annotation refinement (IAR) procedure used to improve CNN performance through classifier-assisted resolution of inter-annotator inconsistencies on the training data. }\label{fig:neural_net}
\end{figure}

\subsubsection*{CNN classifier}
 The CNN architecture is presented in Fig. \ref{fig:neural_net}a. The system consists of three key stages: the first  ``sensor module'' is responsible for low-level feature extraction of individual sensors, and is inspired by the multi-IMU sensor CNN architecture\cite{Ha2016} by having independent 2D convolution kernel paths for the accelerometer and gyroscope signals and one shared kernel that spans both. The ``sensor fusion module'' is responsible for fusing individual sensor-level features to common high-level features. Finally, the ``time series modeling module'' performs temporal modeling of the learned high-level features in order to utilize temporal contextual information in the classification decisions. Time series module was implemented with a residual network architecture \cite{Kaiming2015} utilizing stacks of dilated convolutions over the frame-level features \cite{vanDenOord2016}. Detailed description of the CNN classifier and the training procedure can be found from the supplementary material.

The CNN classifiers for the posture and movement tracks were otherwise identical, but the movement track uses an additional one-hot conditioning vector of the posture classification output (arg max) at the input of the time series modeling module. This addition is justified by the fact that the movement categories are heavily conditional to the postural context (e.g., commando crawling does not happen during the supine posture), and hence information from the posture classification helps to disambiguate sensory signals in the movement classification task. 

\subsubsection*{Iterative annotation refinement for automatic classifier training}
\label{sec:label_refinement}

 Performance of supervised machine learning classifiers is highly dependent on the consistency of the labels used in the training. In our present dataset, there was an incomplete agreement among the three experts in about one third of the frames in the movement track, which is a typical situation in annotation tasks of this kind \cite{Danker-hopfe2009,Massey2019}. In order to utilize all available information, including the observed lack of full agreement in specific frames, we developed a novel iterative annotation refinement (IAR) method. It aims to resolve ambiguities in the training data by combining human- and machine-generated labels in a probabilistic fashion (Fig. \ref{fig:neural_net}b).

In IAR, the data from \begin{math}N\end{math} = 22 infants were first split into a \emph{training set} of \begin{math}N-1\end{math} infants and a \emph{held-out test} infant. For each infant in the training set, the labels from the three parallel annotators were treated as probabilistic priors for class identity for each signal frame at time \begin{math}t\end{math}, i.e., each class  \begin{math}c\end{math} received a prior probability \begin{math}p_0(c | t)\end{math} of 0, 0.33, 0.67, or 1  based on 0/3, 1/3, 2/3, or 3/3 human ratings for the given class.
During the first iteration \begin{math}i\end{math}, an automatic classifier with parameters \begin{math}\theta_i\end{math} was trained on \begin{math}N-2\end{math} of the infants in the training set using the prior probabilities as soft labels. The resulting classifier was then used to estimate new class likelihoods \begin{math}p(c | t, \theta_i)\end{math} for the \begin{math}N-1\end{math}:th infant in the training set, repeating the process \begin{math}N-1\end{math} times to get the likelihoods for each infant in the training set.
The resulting machine ``annotations'' were then combined with the original human annotations through multiplication, i.e., \begin{math}p_i(c | t) = p_0(c | t)*p(c | t, \theta_i) / \sum_c \end{math}, where $\sum_c$ is a normalization constant making each frame's probabilities sum up to 1. The resulting \begin{math}p_i\end{math} were then treated as
soft labels for training a full classifier from all \begin{math}N-1\end{math} infants in the training set to be used for performance evaluation on held-out data, and as new priors (soft labels) for label refinement on the next iteration round. On each iteration, classifier-based likelihoods were combined with the original human labels \begin{math}p_0(c | t)\end{math} (instead of the priors from the previous iteration) in order to ensure that the IAR did not diverge from the original set of classes proposed by the humans.

As a result, the unambiguous frames with 3/3 and 0/3 human category decisions remained unchanged, whereas the ambiguous cases between the two or three competing classes of ambiguous segments were allowed to change based on the classifier decisions. Since the classifier relies on systematic global structure available in the training data, this process leads to unification of the labels in terms of class identities and their corresponding temporal boundaries. In other words, IAR reduces the label noise in the training data, thereby boosting the performance of the resulting final classifier.

\subsection*{Experimental setup and evaluation}
Performance of both classifiers was evaluated using leave-one-subject-out (LOSO) cross-validation, and where IAR with five iterations was used to preprocess the annotations of the subjects in the training set. Notably, system performance was always and only evaluated against the original human annotations, and the IAR was never applied to the test set samples of held-out infants. The SVM classifier was also used to test the effect of various sensor configurations on classification performance, including cases with movement recording from single arm, single leg, two arms, two legs, and all limbs, respectively.

\begin{figure}[t]
\centering
\includegraphics[width=0.9\linewidth]{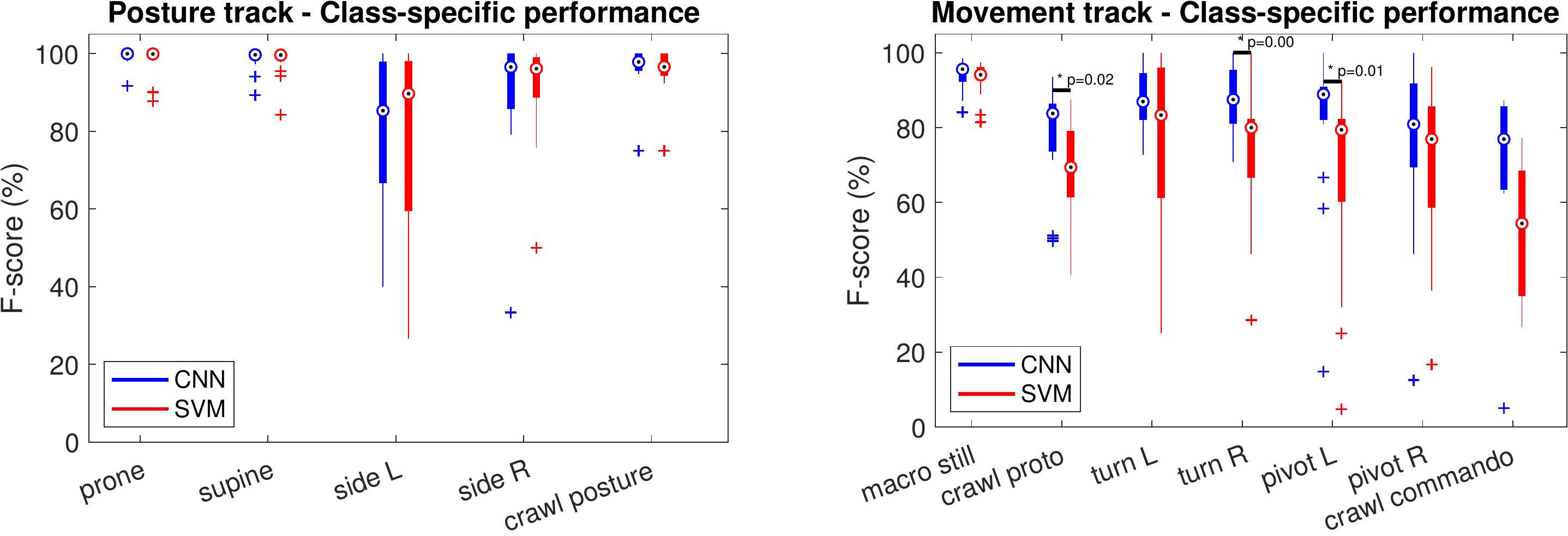}
\caption{Performance of classifiers. Class-specific F-score box plots for individual recordings for the Posture and Movement tracks using the CNN (blue) and SVM (red) classifiers. Statistically significant recording-level differences (p < 0.05; Mann-Whitney U-test, N=22) between SVM and CNN performance are indicated. Note the significantly better performance of CNN in movement patterns that take place in prone position (crawl proto, turn, pivot). }
\label{fig:ab_f1}
\end{figure}

We report the classification performance across all infants (LOSO-folds) and using two sets of test frames: the ``full agreement'' set, which contains the frames from each recording with full 3/3 annotation agreement, and the ``all frames'' set, which contains all of the dataset frames with ground truth based on class majority vote. The ``full agreement'' set is used to showcase classifier performance on unambiguous cases (as judged by humans), and the ``all frames'' set to showcase the overall performance. The differentiation of the two test cases aims to bring better insight on the classification performance between the cases where the ground truth is well defined (the ``full agreement'' set), and the cases that include noisy ground truth labels (the ``all frames'' set). The ``full agreement'' set is used as the primary benchmark in Figures \ref{fig:ab_f1}, \ref{fig:multi_sensor}, and \ref{fig:baby_comparison}.

Four primary performance metrics were used: the accuracy, unweighted average recall (UAR), precision (UAP), and F-score (UAF). For each class, recall is defined as R = tp/(tp+fn) and precision as P = tp/(tp+fp), where tp is the number of true positives, fp the number of false positives, and fn the number of false negatives. The F-score is the harmonic mean of precision and recall, F=2PR/(P+R). The unweighted average score means that the class-specific scores are first calculated and the averaged across the classes, thereby ignoring the highly skewed class distribution of the present data. Thus, the chance-level performance for all of the unweighted average scores is 1/\begin{math}C\end{math}, where \begin{math}C\end{math} is the total number of posture/movement classes. Any statistically significant differences in the results are reported using the Mann-Whitney U-test \cite{mann1947} with $p<0.05$ criterion for significance.

In addition to the primary metrics above, pairwise $\kappa$s between the classifier and each of the human annotators were calculated in order to investigate whether the machine can achieve human-like consistency in the posture and movement classification tasks.

\section*{Results for automatic classification}

\subsection*{Assessment of classifier performance}
The result comparisons for SVM and CNN classifiers are shown in Fig. \ref{fig:ab_f1} for all recordings in terms of their category-specific F-scores. The classification accuracies of posture are generally comparable between CNN and SVM, however the CNN classifier yields a notably better performance with several movement categories. The median F-score for the movement track is approximately 80\% for the CNN classifier, with the SVM being consistently around 5 to 10 percentage points below in performance. With respect to the  intended use of the smart jumpsuit in clinical assessment, the greatest shortcomings with the SVM classifier are its significantly worse performance in classifying movements that take place in prone posture: \texttt{crawl commando}, \texttt{crawl proto} and \texttt{pivoting}.

\begin{figure}[th]
    \centering
    \begin{subfigure}[c]{0.42\linewidth}
        \centering
        \includegraphics[width=0.75\linewidth]{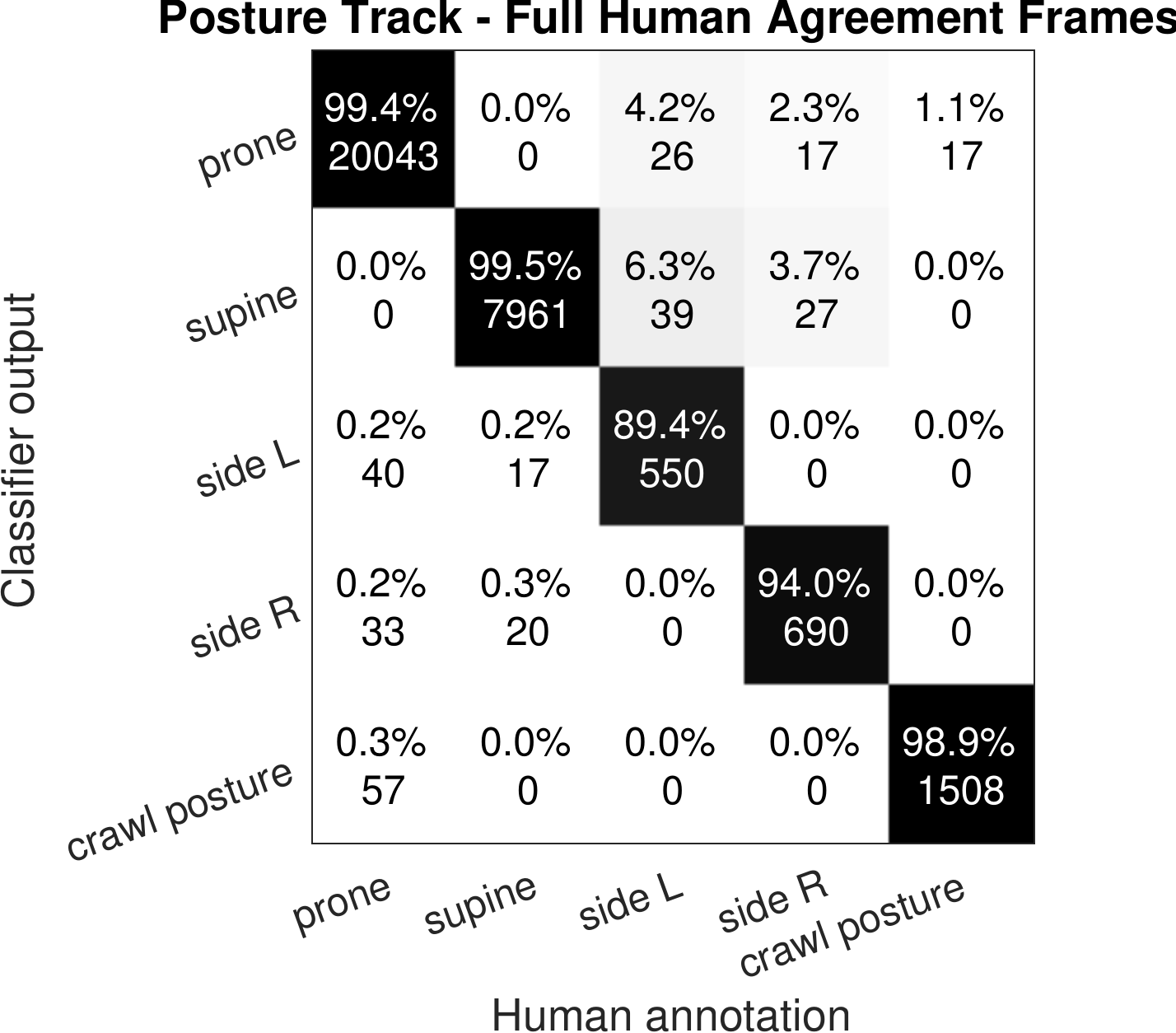}
        \caption{}
        \label{fig:a_test_results}
    \end{subfigure}
    \qquad 
    \begin{subfigure}[c]{0.42\linewidth}
        \centering
        \includegraphics[width=0.75\linewidth]{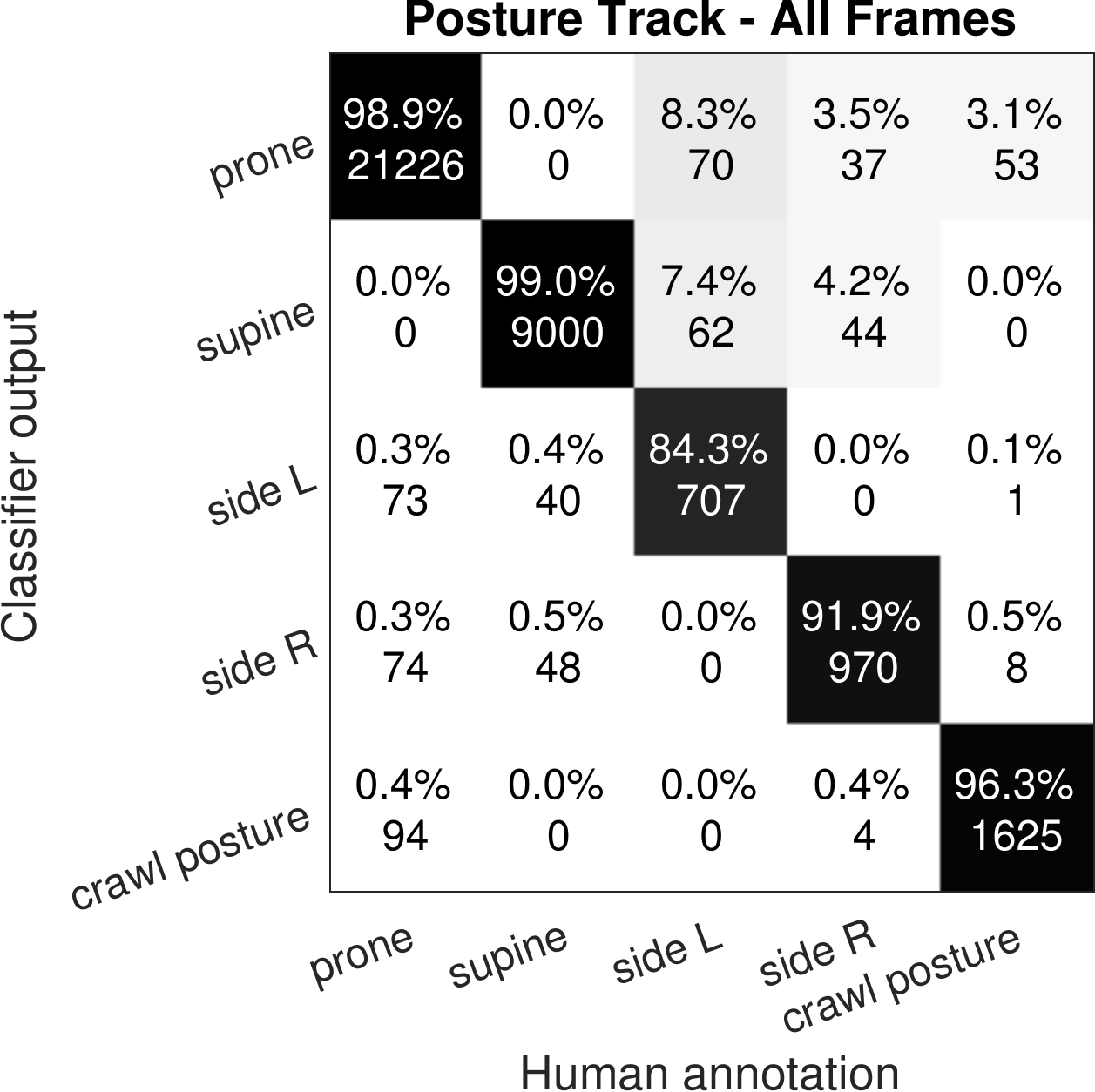}
        \caption{}
        \label{fig:a_all_results}
    \end{subfigure}
    \\ ~ 
    \begin{subfigure}[b]{0.42\textwidth}
        \centering
        \includegraphics[width=\textwidth]{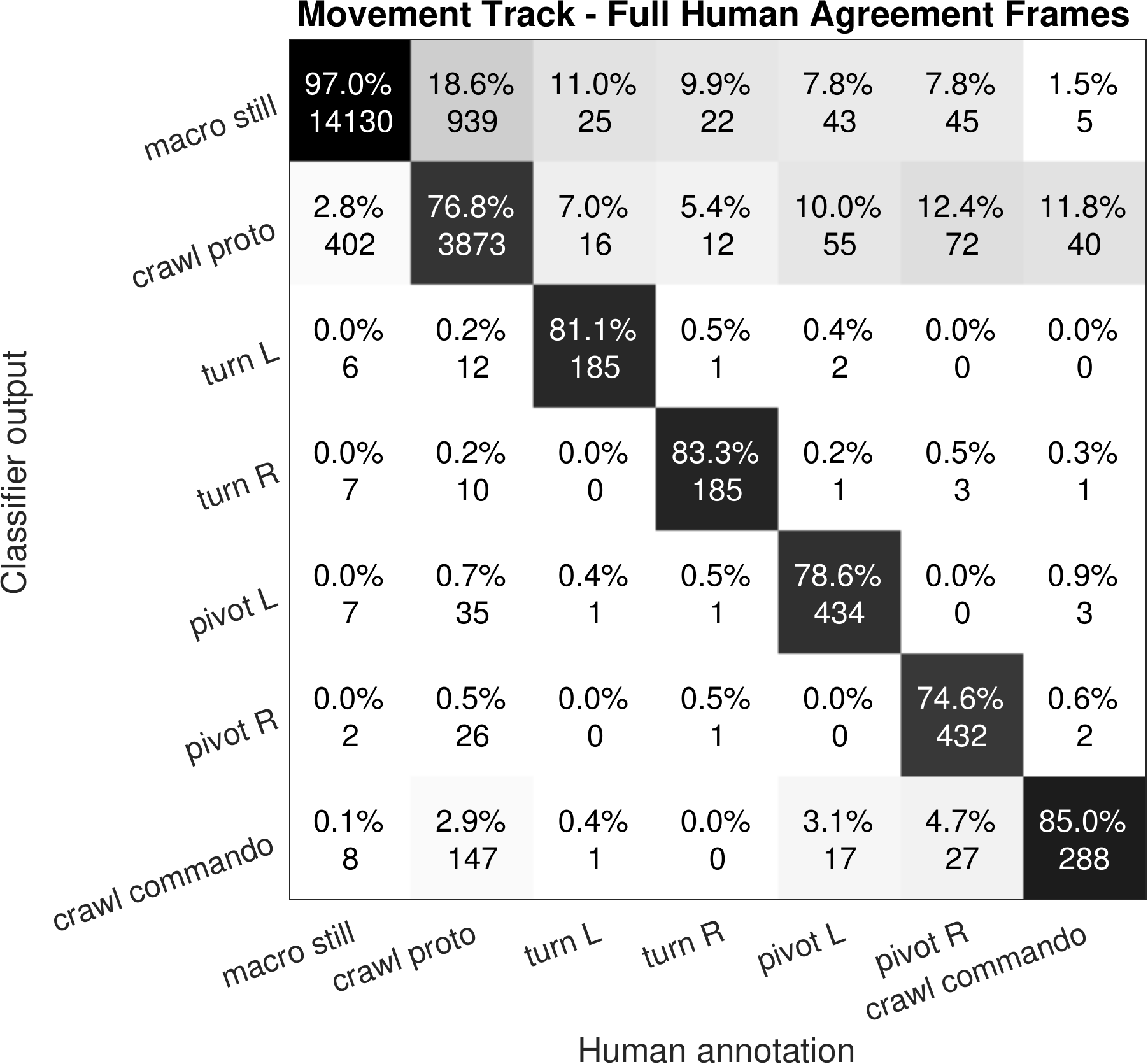}
        \caption{}
        \label{fig:b_test_results}
    \end{subfigure}
    \qquad 
    \begin{subfigure}[b]{0.42\textwidth}
            \centering
        \includegraphics[width=\textwidth]{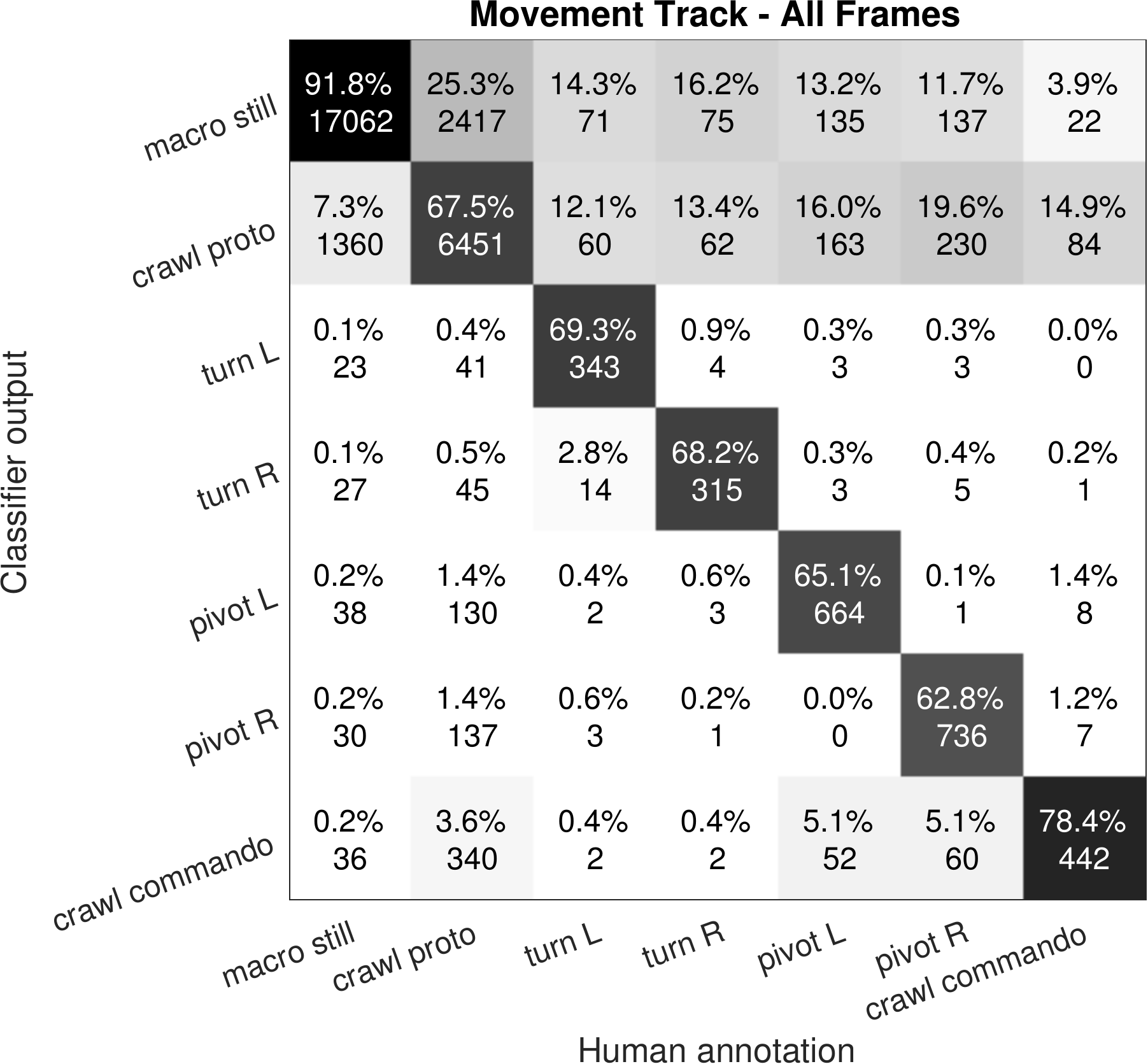}
        \caption{}
        \label{fig:b_all_results}
    \end{subfigure}
    \caption{Total CNN classifier confusion matrices of the posture (a and b) and movement (c and d) tracks obtained from LOSO cross-validation of the (a and c) full annotation agreement subset and (b and d) complete data set. The percentage values inside the cells indicate the class-specific recall values, and the absolute values denote the number of frames. Average metrics are presented in Table \ref{tab:results}.}\label{fig:a_results}
\end{figure}

\begin{table}[ht]
\caption{Overall performance of the CNN classifier in terms of accuracy (ACC), unweighted average recall (UAR), unweighted average precision (UAP), and unweighted average F-score (UAF).} \label{tab:results}
\begin{tabular}{|l|llll|}
\hline
\textbf{Full agreement frames} & \textbf{ACC} & \textbf{UAR} & \textbf{UAP} & \textbf{UAF} \\ \hline
Posture track                  & 99.1\%       & 96.2\%       & 95.7\%       & 96.0\%       \\ 
Movement track                 & 90.7\%       & 82.4\%       & 85.9\%       & 83.5\%       \\ \hline \hline
\textbf{All frames}            &  \textbf{ACC} & \textbf{UAR} & \textbf{UAP} & \textbf{UAF}      \\ \hline
Posture track                  & 98.2\%       & 94.1\%       & 93.3\%       & 93.7\%       \\ 
Movement track                 & 81.7\%       & 71.9\%       & 75.4\%       & 72.7\%       \\  \hline
\end{tabular}
\end{table}

The confusion matrices in Fig. \ref{fig:a_results} show class-specific performance metrics for the CNN classifier, and the overall performance scores in terms of accuracy, UAR, UAP, and UAF are reported in Table \ref{tab:results} (for more details, see Supplementary material). The results are shown separately for the full annotator agreement frames (the frames whose ground truth labels are most likely absolutely correct), and for all frames (contains the 3/3 and the ambiguous frames with noisy labeling). As can be observed, the confusion matrices are highly similar to the confusion matrices of human annotation presented in Figure \ref{fig:annotaion_results}a. For posture, overall performance is highly similar between the ``full agreement'' and ``all frames'' sets, which reflects the high level of inter-rater agreement. The most frequent confusions are in the prone-side-supine axis, as well as between crawl posture and prone. Notably, confusion in the left-right axis does not exist with automatic classification but they may occur by human errors. The movement track confusion matrix from the classifiers (Fig. \ref{fig:a_results}c and d) is also strikingly comparable to the confusion matrix between human experts (\ref{fig:annotaion_results}a). The most obvious challenge is at the decision boundaries between \texttt{macro still}, \texttt{crawl proto}, and the rest of the categories.

\subsection*{Assessment of IAR effectiveness on annotation consistency and classifier performance}
\begin{figure}[ht]
\centering
\includegraphics[width=0.9\linewidth]{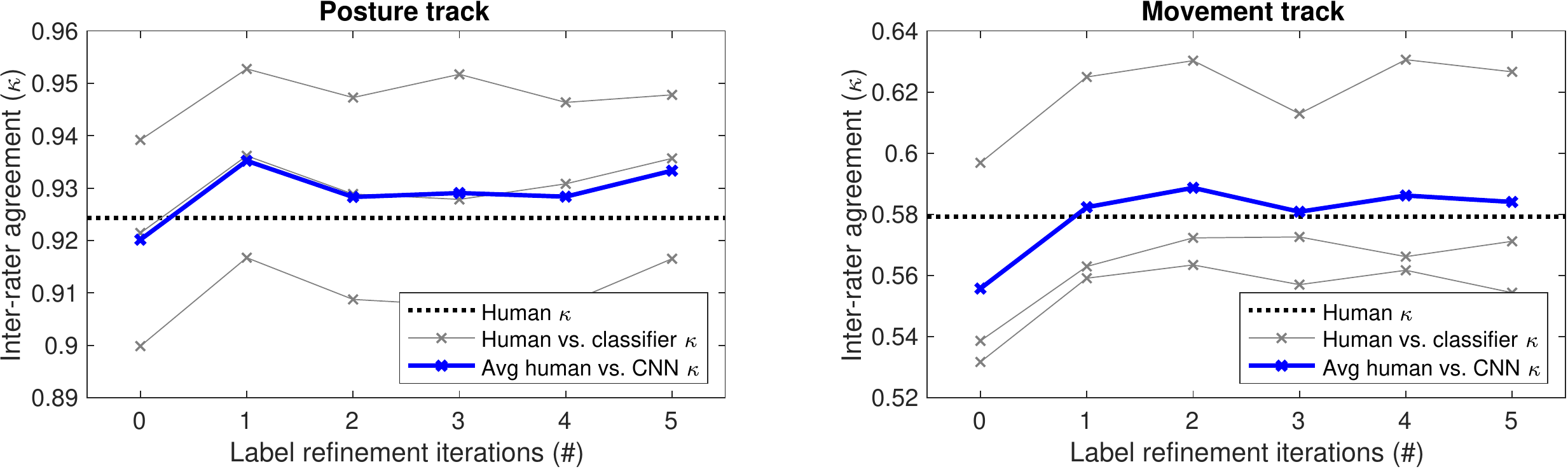}
\caption{The effect of IAR-based label refinement on the agreement ($\kappa$) between CNN classifier outputs and the three original human annotation tracks (gray lines) as a function of IAR iterations on held-out test data. The mean pair-wise agreement between classifier and the three human annotations is shown with the blue line and Fleiss' $\kappa$ agreement across all human annotators is shown with the dotted black line. A clear improvement in classifier performance is observed due to label refinement on the training data, reaching the human-to-human  agreement rate.}
\label{fig:label_refinement_kappa}
\end{figure}

Figure \ref{fig:label_refinement_kappa} shows the value of the IAR procedure with the CNN classifier, as measured in terms of Fleiss' $\kappa$ between human annotators and the classifier outputs on held-out test data as a function of IAR iterations applied to the training data. Classifier performance saturates after the first, notable increase over the first IAR iterations, indicating that the classifier has learned more consistent decision boundaries. In addition, the saturating performance demonstrates that the proposed probabilistic framework with the internal cross-fold procedure within the IAR (not to be confused with the LOSO-loop of the main experiment) ensures relative stability of the approach.

The results in Figure \ref{fig:label_refinement_kappa} also show how the classifier behavior compares with different human annotation tracks. Comparison of human-to-human agreement (dashed lines in Fig. \ref{fig:label_refinement_kappa}) with the classifier-to-human agreement showed that a human annotator can be replaced by our classifier without loss of agreement, and the finding holds for both posture and movement tracks. A high comparability of human and computer assessment is even seen in the level of individual participants (see Fig. \ref{fig:baby_comparison}). Taken together, the results show that the classifier performs at human equivalent level.

\begin{figure}[ht]
\centering
\includegraphics[width=0.4\linewidth]{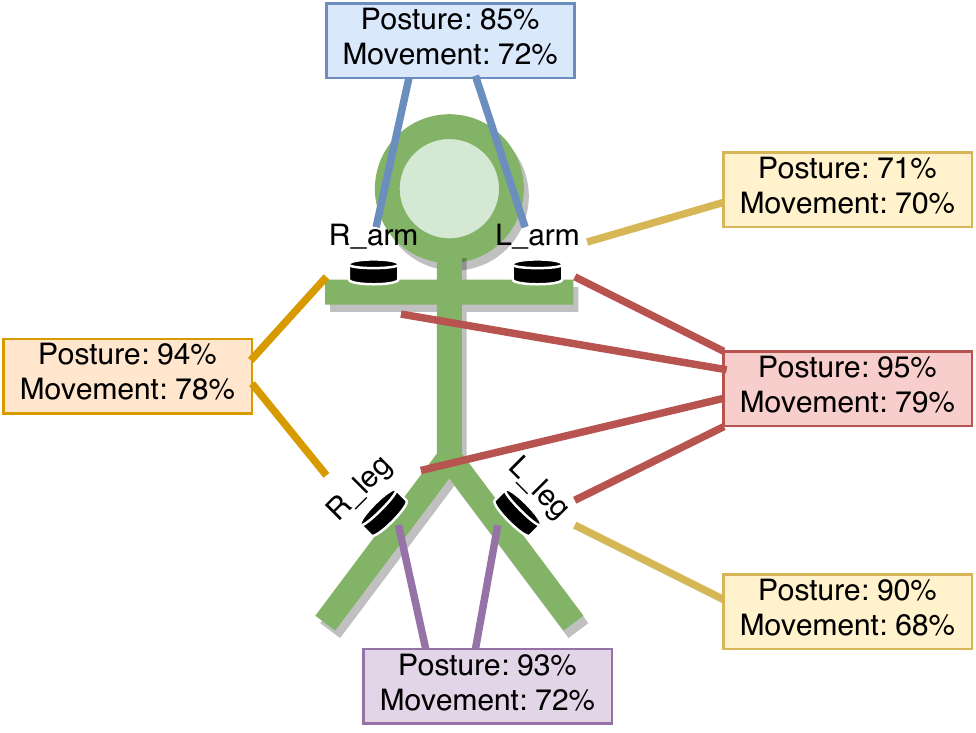}
\caption{The effect of sensor setup on classification performance (UAR) using the SVM classifier. Any individual sensor configuration is inferior to the four-sensor setup. However, classification of data from a combination of one arm and one leg leads to almost comparable results.}
\label{fig:multi_sensor}
\end{figure}
\subsection*{Added value from multi-sensor setup}
In order to investigate the added value of multiple sensors, we compared classifier performance (UAR) using different subsets of sensor configurations (Fig. \ref{fig:multi_sensor}). The main finding was that the individual sensors show lower performance compared to the full four-sensor setup. A two-sensor combination with one arm and one leg may lead to a performance that is comparable to the four-sensor setup in the given task. However, such two-sensor setup makes the recording more susceptible to problems that may arise from sensor placement errors or hardware failures. In addition, recordings with only one leg or arm cannot measure laterality differences of limb synchronies which may be central in future development of actual automatic diagnostic paradigms \cite{Guzzetta2003} (see also Discussion).

\subsection*{Automatic recognition of infants with different levels of motor performance} 
\begin{figure}[ht]
\centering
\includegraphics[width=0.9\linewidth]{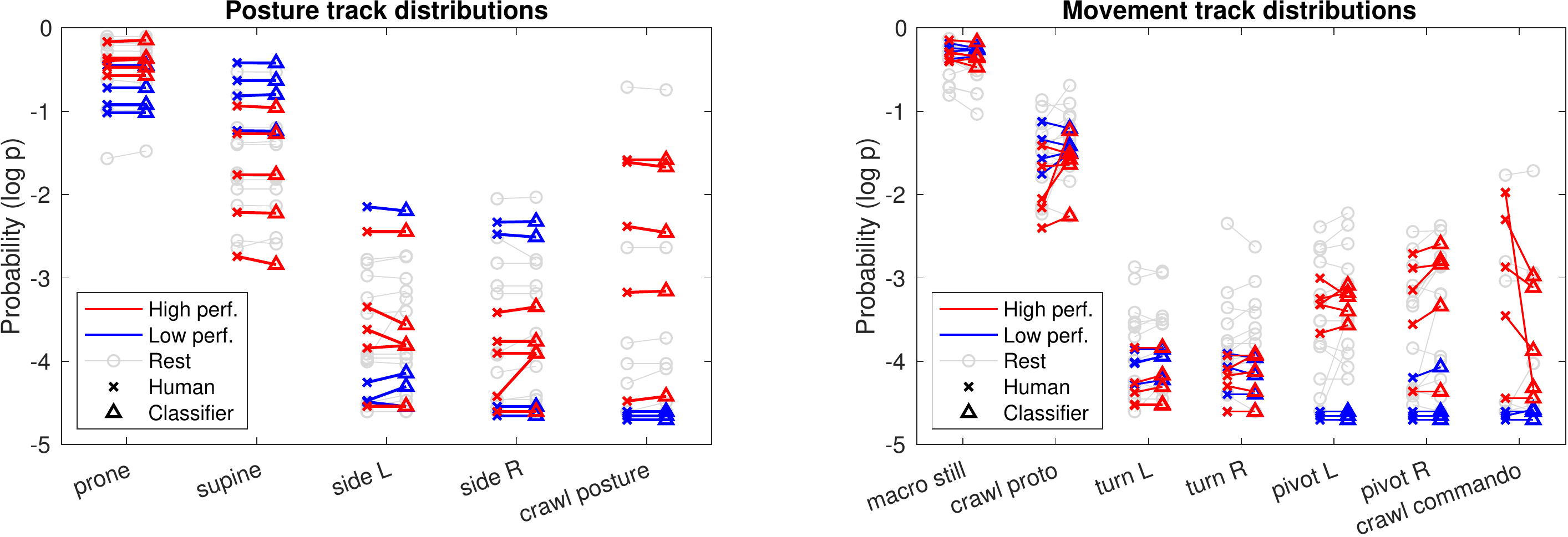}
\caption{Differentiation of high and low motor performance infants with the smart jumpsuit. The plots show individual category distributions (as log-probability) of the given posture and movement category, presented for the entire dataset. Results from both human annotation (x) and the classifier output (triangle) are shown for comparison, and the hairlines connect individuals to assess the individual level reliability. The highlighted recordings correspond to a sample of high performing (red; High perf.) and low performing infants (blue; Low perf.). Rest of the infant cohort is plotted with light gray lines. No statistically significant differences were found between the movement distributions from the human annotations and from classifier outputs (Mann-Whitney U-test, $N=22$).  }
\label{fig:baby_comparison}
\end{figure}

We finally carried out a proof-of-concept test to see if the current jumpsuit setup is able to automatically distinguish infants with high versus low motor performance. To this end, five infants with high and five with comparatively low motor performance levels were chosen from the video recordings by a professional child physiotherapist (T.H.) and a child neurologist (L.H.). The recordings were chosen based on a separate session of retrospective video review with a consensus assessment approach. High performance was defined as a generally active motility and rich movement repertoire, or relative paucity of both in the low performing infants, respectively. The relative frequency of all posture and movement events was then plotted per category to qualitatively assess whether differences in performance would be distinguishable in the output of the automatic analysis (Fig. \ref{fig:baby_comparison}). Indeed, this analysis shows that the two infant groups are far apart in the incidence of several categories in both posture (\texttt{prone}, \texttt{supine}, \texttt{crawl posture}) and movement (\texttt{macro still}, \texttt{pivot L} and \texttt{R}, \texttt{crawl commando}) tracks. In addition, the two groups are clearly separable in terms of both human and machine -based labels.

\section*{Discussion}

This study shows that it is possible to construct a comfortable-to-wear intelligent infant wearable with a signal processing pipeline that allows quantitative tracking of independent movement activities of infants with high accuracy. We developed a novel annotation scheme to classify infant postures and movements into a number of key categories, and demonstrated how an automatic classifier can reach human-like consistency in movement and posture recognition. In addition, we described a principled probabilistic approach to exploit the inter-rater inconsistencies in the human annotations used to train the classifier. Finally, we demonstrated that a multi-sensor setup is required for maximal movement classification performance. Our present work extends the prior work from adult studies \cite{Ha2016} that clinically relevant movement tracking and quantification is possible in infants as well. The present work goes beyond the prior literature by constructing and demonstrating the feasibility of the first multi-sensor wearable for infants that allows non-intrusive, cheap, and technically achievable measurement of infants' posture and independent movements. The result is a full smart jumpsuit system that could be implemented in out-of-hospital recordings, at least in the clinical research context.

From the technical point of view, our findings show that an SVM classifier based on standard signal-level features is sufficient for posture tracking and adequate for detecting some categories of movement. However, the SVM struggles in recognizing certain key infant motor patterns, such as crawling posture and pivoting, which are crucial milestones for a normal neurological development \cite{Sharma2011}. 
For this reason, an end-to-end CNN classifier was designed for the task, as similar CNN architectures have previously demonstrated state-of-the-art performance in adult-based human activity recognition \cite{Ha2016}. The resulting CNN classifier yielded statistically significant and much needed/required improvements in the movement tracking. Comparison to human annotator agreement levels shows that the CNN achieves classification performance comparable to human annotator consistency, and it could therefore be used as an independent automatic measure of infant motility.

\subsection*{Practical aspects of jumpsuit analysis}
Literature on movement analysis based on IMU sensors has grown rapidly and a wide range of analytic tools have been developed to analyse movement activity at different levels \cite{Clark2017, Migueles2017}. 
A key challenge has been in search for unified recording settings and/or classification tasks \cite{Clark2018}. 
It now seems clear that solutions need to be tailored specifically, at least for different subject groups and tasks \cite{Migueles2017}. 
For instance, there may be substantial day-to-day variation in motility, which needs to define a balance between the added information gain vs practical costs of longer recording periods \cite{Ricardo2018}. 
Preliminary studies have shown, however, that quantitative movement analysis of infants may be possible with accuracy that even allows clinical outcome predictions \cite{Abrishami2019, Lobo2019, Goodfellow2018}. 

Designing an infant medical wearable of this kind is a multidisciplinary challenge \cite{Lobo2019}. At the patient level, there are practical challenges such as wearing comfort to allow normal motility. Here, we chose an infant swim suit as the model for cut design;  the sensor placements were such that they would be likely ignored by most infants to allow undisturbed motility. We initially also piloted with distal sensor placements, but they were not found to bring significant benefits for automated analysis. Instead, distal sensors tended to be mechanically more unstable and easily attracted infants' attention.

At the level of operator, the full recording system including the mobile device for data collection needs to be easy enough to use, while the collection of synchronized data must be reliable throughout the session. A further improvement of the system could be achieved by development of higher memory capacity into the sensor modules, thereby lifting the need for continuous data streaming. An overarching issue is the need to reduce complexity of the whole setup. Our smart jumpsuit design was markedly challenged by the chosen multi-sensor setup, which required reliable wireless collection of synchronized data at high rates over Bluetooth transmission. The choice to use multiple sensors was intuitively reasoned by a potential for better movement discrimination in follow-up research. Comparison of classification results between different sensor configurations shows that posture detection alone may be relatively reliable from even one sensor. However additional sensors bring substantially more accuracy to recognition of movement patterns. In the context of prospected out-of-hospital studies, it is also important to consider technical reliability, including participant's compliance and hardware-related issues. Any additional data, such as redundant sensor information, may prove invaluable in the real life settings of infants' native environment (see also \cite{Lobo2019, Block2016}).

\subsection*{Future prospects}
Multisensor recording of the present kind opens many potential ways for further analyses. For instance, one can readily envision quantitative, posture context-dependent assessment, where computational analyses account for different postures. For instance, recent literature reports computational measures of spontaneous hand or leg movements that are highly predictive of later neurodevelopment \cite{Abrishami2019,Harbourne2015}. Infants' intentional hand or leg movements are heavily modulated by posture, yet no method is available to allow their analysis separately for different posture contexts; an approach that becomes readily available with the methodology described in out present work.
Moreover, posture and movement tracking enables context-specific analysis of heuristic features (e.g., \cite{Goodfellow2018,Torres2016,Trujillo-Priego2017}) such as movement symmetry and limb synchrony, the well established metrics in the analysis of adults' spontaneous motility \cite{Knaier2019}.

Recent literature has underscored the need to develop functional growth charts to allow evidence-based tracking of individual neurodevelopment \cite{Lobo2019,Torres2016}; On another note, recent changes in infant care practices have emphasized ``tummy time'', placing infants on their stomach to play when awake, as an important posture context to support early neurodevelopment \cite{Majnemer2005}. All these trends will benefit from an established, automated methodology that allows tracking posture and movement patterns. As an initial proof of concept, we showed how posture and movement tracking may differentiate high and low performing infants by simply quantifying mere incidence of movement categories. Further efforts with context-dependent quantitation are likely to boost the information value. An obvious practical use case of such method would be tracing atypical patterns in motor development. Moreover, a reliable quantitative tool for motor activity tracking holds significant promise for a functional biomarker, i.e., providing much awaited evidence for the efficacy of early therapeutic interventions \cite{Herskind2015,Novak2017}.

\bibliographystyle{abbrv}
\bibliography{refs}


\section*{Acknowledgements}

The research was funded by Academy of Finland grants no. 314602 and 314573, as well as project grants from Pediatric Research Foundation, Aivos\"{a}\"{a}ti\"{o} and Juselius foundation.

\section*{Author contributions statement}

M.A., S.V., L.H., E.I., O.R., T.H., and V.M. conceived the study. S.V., O.R., and L.H. supervised the work. M.A. executed the computational components of the study; N.K. developed the mobile application; T.H. and M.A. executed the experimental aims and M.A., A.K., S.B., A.V., and A.G. performed human annotations. All authors reviewed and approved the manuscript.

\section*{Additional information}

\textbf{Competing interests}: The authors declare no competing interests.



\end{document}